\documentclass[10pt,twocolumn,letterpaper]{article}

\usepackage{cvpr}              

\usepackage{graphicx}
\usepackage{amsmath}
\usepackage{amssymb}
\usepackage{booktabs}
\usepackage[pagebackref,breaklinks,colorlinks]{hyperref}

\usepackage[capitalize]{cleveref}
\crefname{section}{Sec.}{Secs.}
\Crefname{section}{Section}{Sections}
\Crefname{table}{Table}{Tables}
\crefname{table}{Tab.}{Tabs.}

\def\cvprPaperID{*****} 
\def\confName{ECCV}
\def\confYear{2022}

\begin{document}

\title{Rooms with Text: A Dataset for Overlaying Text Detection}

\author{Oleg Smirnov\\
Amazon\\
{\tt\small osmirnov@amazon.com}
\and
Aditya Tewari\\
Amazon\\
{\tt\small aditewar@amazon.com}
}
\maketitle

\begin{abstract}
  In this paper, we introduce a new dataset of room interior pictures with
  overlaying and scene text, totalling to 4836 annotated images in 25 product
  categories. We provide details on the collection and annotation process of
  our dataset, and analyze its statistics. Furthermore, we propose a baseline
  method for overlaying text detection, that leverages the character
  region-aware text detection framework~\cite{baek2019character} to guide the
  classification model. We validate our approach and show its efficiency in
  terms of binary classification metrics, reaching the final performance of
  0.95 F1 score, with false positive and false negative rates of 0.02 and 0.06
  correspondingly.
\end{abstract}

\section{Introduction}
\label{sec:intro}

Text detection and recognition task in images or videos typically do not
distinguish between overlaying and scene text. Overlaying text is artificially
superimposed on the image at the time of editing, and scene text is a text
captured by the recording system. Distinguishing the overlaying text from the
scene text is an important computer vision problem with multiple practical
applications.

In video processing, locating and labeling superimposed captions is a pre-text
stage for semantic video indexing, summarization, video surveillance and
security, and multilingual video information access. In the e-commerce and
social media domains, detecting overlaying text is paramount for identifying
fraudulent and forbidden content, such as spam and unsolicited advertising in
user generated imagery. Overlaying text recognition also finds important
applications in forensic science~\cite{bhardwaj2016image}.

The problem is particularly prominent in scenes with complex background, where
one can expect large amounts of organic text. Those examples include but are not
limited to urban street landscapes with outdoor signs, room interiors with
wall decor, books in shelves, clocks, and similar objects.

Despite of recent progress in robust text in scene detection, state-of-the-art
approaches are not applicable to the problem at hand, since conventional detectors
typically specialize on one kind of text only. From the practical point of
view, non-specialized models can not be deployed for fraudulent and forbidden
content detection scenarios, which are highly sensitive to false negative
rates.

In this paper, we present an annotated database that consists of real-world
e-commerce images with overlaying and scene text in various combinations. We
describe the collection and annotation process of our dataset, and analyze its
statistics. Further, we propose a baseline method for overlaying text
identification, which incorporates text localization to attend to relevant
image regions for classification. We benchmark the performance of our method
against an off-the-shelf logo and watermark detection algorithm to verify its
effectiveness.

The rest of the paper is organized as follows. In Section~\ref{sec:related}
we review relevant work in the field. In Section~\ref{sec:dataset} we
introduce and analyze the Rooms with Text dataset consisting of room interior
images annotated with text type labels. Further, in Section~\ref{sec:method}
we propose a simple, yet competitive baseline for distinguishing overlaying
text from scene text in images. In Section~\ref{sec:experiments} we discuss
results of experiments and ablation studies. We conclude and outline future
work in Section~\ref{sec:conclusions}.

\section{Related Work}
\label{sec:related}
Most of the literature in overlaying text detection is motivated by the task
of recognizing video subtitles, also known as captions in video streaming.

Conventional methods operate under assumption that superimposed text regions
have distinctive visual characteristic that can be identified by analyzing
color~\cite{zhang2003accurate} and texture
information~\cite{kim2008new}. Edge-based approaches are also considered
useful since text areas contain rich edge
information. \cite{shivakumara2014separation} proposed separating the
instances based on the distribution of outputs produced by Canny and Sobel
edge detectors. \cite{quehl2015improving} trained a classifier on a linear
combination of the histogram of oriented gradients
features. \cite{kumari2018three} developed a multi-resolution algorithm that
identifies regions of interest on a color histogram curve, followed by
connected component analysis on candidate text areas.

Another stream of work studied regions of high contrast that can be
characterized by saliency methods~\cite{lienhart2002localizing}, as well as
more subtle features like high frequencies determined from wavelet
coefficients~\cite{gllavata2004text}, and compression algorithm
artefacts~\cite{bhardwaj2016image}. Among deep learning models,
\cite{slucki2018extracting} proposed an end-to-end pipeline that combines
TextBoxes~\cite{liao2017textboxes} text detection algorithm with a
convolutional recurrent neural network for extracting overlays from social
media videos.

However, video subtitles recognition constitutes a relatively simple subset of
a more generic overlaying text detection task. Video overlays are located in a
designated position in the image, have strictly horizontal orientation, and
often depicted on a contrasting background to optimize for readability.
Moreover, above mentioned approaches are commonly evaluated on synthetic
data. In absence of standardized benchmarks and labeled datasets, it is
challenging to compare and evaluate their performance for real life
applications.

Besides the direct methods, the task of overlaying text detection can be
tackled by re-purposing approaches in logo and watermark detection, and generic
scene text recognition.

\textbf{Watermark and logo detection} from images has been extensively studied
in computer vision and pattern recognition literature. To this end, a
watermark is understood as a semi-transparent logo or inscription that is
added to the image at post-processing stages with a purpose of intellectual
property protection, product brand management on social media, and others.

The task of watermark detection is similar to overlaying text detection in the
sense of identifying image regions that do not organically belong to the
scene. However, it is different in that image watermarks make use of
semi-transparent alpha compositing, that enables usage of strong image
priors~\cite{ulyanov2018deep} for detection and removal.

Identifying and removing a watermark from a single image without a-priori
information is a daunting task. Common approaches frame it as a multi-image
problem, that assumes that watermarks are added in a consistent manner to many
images. Classical methods leverage gradient information
\cite{yan2005automatic,wang2007automatic} and iterative subsequent
matching~\cite{dashti2015video} algorithm. More recently,
\cite{dekel2017effectiveness,shen2021large} proposed consistency-based
approaches for discovering repeated patterns of low variability in image
collections. SplitNet method~\cite{cun2021split} builds upon a multi-task
network to predict a mask and coarser restored image for further refinement
with a mask-guided spatial attention.

Contrasting to the watermark detection, logo detection rely neither on the
semi-transparency nor on multi-image assumptions. Furthermore, logo detection
formulation assumes availability of a dataset of well-known brand logotypes,
making it a closed-vocabulary problem. Recent families of methods make use of
end-to-end image recognition~\cite{joly2009logo,li2010fast} and object
detection~\cite{hoi2015logo,jain2021logonet} approaches
correspondingly. Research in this field leverages publicly available logo
databases~\cite{romberg2011scalable,hoi2015logo,wang2022logodet} as well as
synthesized datasets~\cite{su2017deep,sage2018logo}.

\textbf{Scene text recognition.} The problem of accurate text recognition and
localization was solved with the abundant availability of datasets such as
MSRA-TD500~\cite{yao2012detecting}, ICDAR~2015~\cite{karatzas2015icdar},
COCO-Text~\cite{veit2016coco}, and Total-text~\cite{ch2017total}. Recent
challenges~\cite{chng2019icdar2019} include text instances arranged in curved
and other irregular shapes, that further contribute to robustness of detection
algorithms in real world scenarios.

Unlike objects in general, texts are often presented in irregular shapes with
various aspect ratios. To handle this problem, EAST~\cite{zhou2017east}
directly predicts geometry maps that combine rotated boxes with quadrangle
coordinates. Another common approach is based on works dealing with
segmentation, which aims to seek text regions at the pixel level. For example,
Pixellink~\cite{deng2018pixellink} detects texts by estimating word bounding
areas. End-to-end approaches, e.g. FOTS~\cite{liu2018fots}, concatenates and
trains the detection and recognition modules simultaneously so as to enhance
detection accuracy by leveraging the recognition result. State-of-the-art
character-level text detectors, such as CRAFT~\cite{baek2019character}
(Character Region Awareness For Text detection), use text block candidates and
produce probability maps to identify individual characters.

\section{Dataset}
\label{sec:dataset}
In this paper, we propose Rooms-with-Text -- a fully annotated dataset of room
interior images with overlaying and scene text. Dataset construction comprises
of three steps, namely candidate image selection, data labeling, and manual
vetting.

\textbf{Candidate image selection.} The dataset contains of publicly available
images collected from a large online retailer. Candidate images were
identified by the means of the CRAFT~\cite{baek2019character} text
detector. CRAFT exhibits an impressive performance in reliably detecting
challenging and indistinct test regions. However, due to the objective of a
character-level approach, the model tends to hallucinate in presence of
letter-like shapes and patterns, which are common in indoor interior design.
As a result, running CRAFT inference without subsequent filtering leads to a
vast amount of false positives. To address this challenge, we devise a
heuristic rule:
\begin{equation}
  G(x) = \frac{4}{h \cdot w}\sum_{\substack{0 \leq i < h / 2\\ 0 \leq j < w /
      2}} F^{\text{rg}}_{i, j}(x) \{ F^{\text{rg}}_{i, j}(x) > T \}
\end{equation}

where $x$ is an input image of size $h \times w$, $F^{\text{rg}} \in
R^{\frac{h}{2} \times \frac{w}{2}}$ is the output region scores matrix of a
trained CRAFT model, and $T = 0.8$ is the region threshold constant proposed
in the paper. We consider images for annotation, if $G(x) > 5 \times 10^{-4}$

We provide sample pictures in Appendix~\ref{sec:dataset_examples}. Notable
examples contain text in various languages and writing systems, including
Chinese characters. Overlaying text is represented by logotypes, inscriptions,
and watermarks with different levels of transparency. Whereas organic scene
text can occur is wall posters, book titles, and other real objects in fonts
of various size. The second from the left image in the first row exemplifies
erroneous bounding boxes produced by the CRAFT algorithm.

\textbf{Data labeling.} Annotation labels were collected from 5 qualified
workers per image, recruited using Amazon Mechanical Turk
platform.\footnote{http://www.mturk.com} Workers' qualification was determined
by the MTurk Masters designation as well as their performance on previous
image labeling tasks.

The annotators were presented with a room image, and asked to categorize it
into 4 non-overlapping classes: ``Overlaying'', ``Organic'', ``Both'', and
``None''. Each vote option was illustrated with 2-3 sample images, manually
curated to be representative of the corresponding class.

We performed 3 rounds of annotation collection, starting with small size data
batches and analyzing label consistency after each stage. We noticed that the
workers often mislabel examples where scene text is hard to notice due to
factors such as tiny font size, visual occlusion, and peripheral location. For
the annotators' convenience, we additionally decorated query images with text
bounding boxes, obtained by the CRAFT text detector.

\textbf{Annotation vetting.} After finishing the previous steps, each image
was manually examined and reviewed to guarantee the quality of data after
selection and annotation. If a labeled example did not meet the quality
requirements, the image was rejected and re-annotated.

By inspecting the distribution of voting times for the pool of annotators in
the initial batch, we observed that the 5\textsuperscript{th} percentile of
voting time is 7 seconds, which we select as a threshold on the minimum time
to consider a label valid. On subsequent stages, we discarded the annotations
corresponding to the lowest 5\% of voting time as possible outliers.

\begin{figure*}[t]
  \centering
  \begin{subfigure}[b]{0.49\textwidth}
    \centering
    \includegraphics[width=\textwidth]{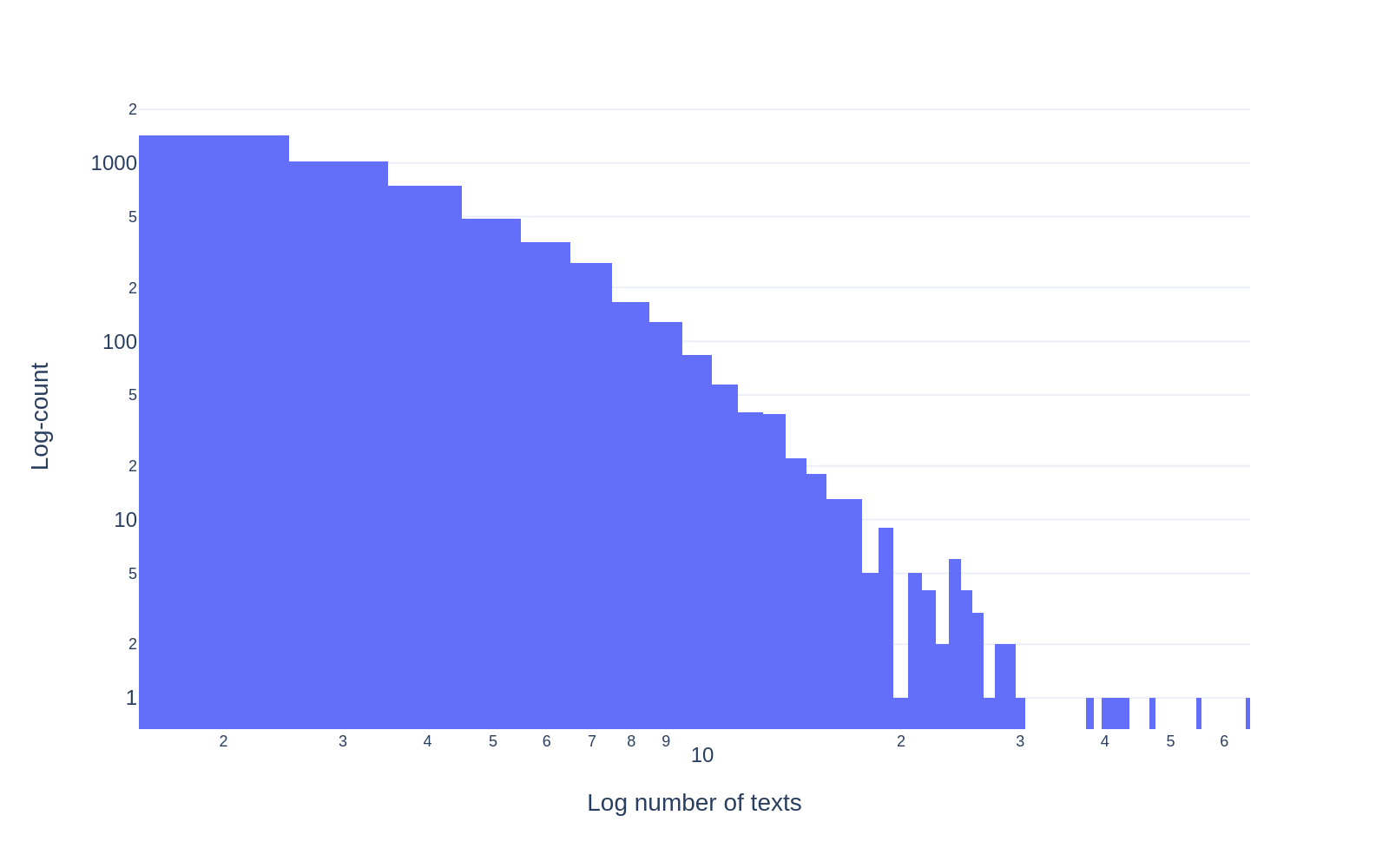}
    \caption{Log-log histogram of the number of~text regions per image}
    \label{fig:regs_distr}
  \end{subfigure}
  \begin{subfigure}[b]{0.49\textwidth}
    \centering
    \includegraphics[width=\textwidth]{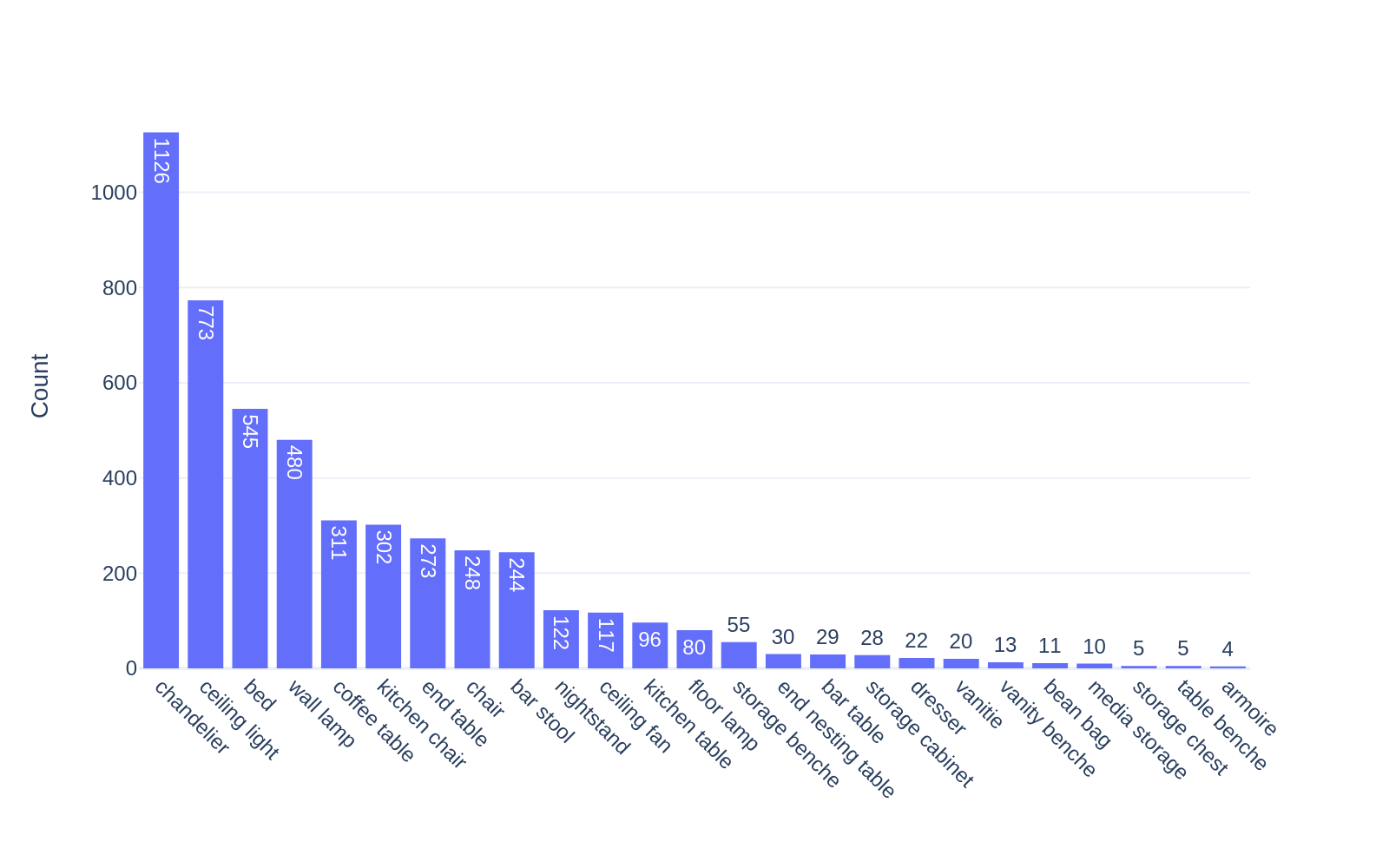}
    \caption{Histogram of the number of image per product category}
    \label{fig:cats_distr}
  \end{subfigure}
  \caption{Rooms-with-Text dataset statistics}
\end{figure*}

\subsection{Dataset Statistics}
The images in the database depict products from 25 Home and Furniture
categories in lifestyle context. Figure~\ref{fig:cats_distr} shows the
distribution of the numbers of images per product
category. Figure~\ref{fig:regs_distr} depicts the distribution of the numbers
of identified text regions on a logarithmic scale. The distribution roughly
follows the Zipf's law.

On the final processing stage, resulting per-example labels were determined by
majority voting across 5 annotations. The histogram of the number of assigned
labels per example is shown in Figure~\ref{fig:regs_distr}. We found that the
overall agreement in votes is fairly consistent across the whole dataset. In
60.2\% of cases all 5 annotators have agreed on the single label. Moreover,
another 34.2\% of examples have 3 or 4 votes for the same label. In total, we
found 65 ambiguous examples with no clear voting majority, that constitutes
1.3\% of the dataset. The labels for those examples were manually reviewed and
corrected during the final cleaning stage.

The final cleaned dataset includes 4836 images, split into 3627 training and
1209 validation examples.

\section{Method}
\label{sec:method}

\begin{figure}[ht]
 \centering
\includegraphics[width=0.8\linewidth]{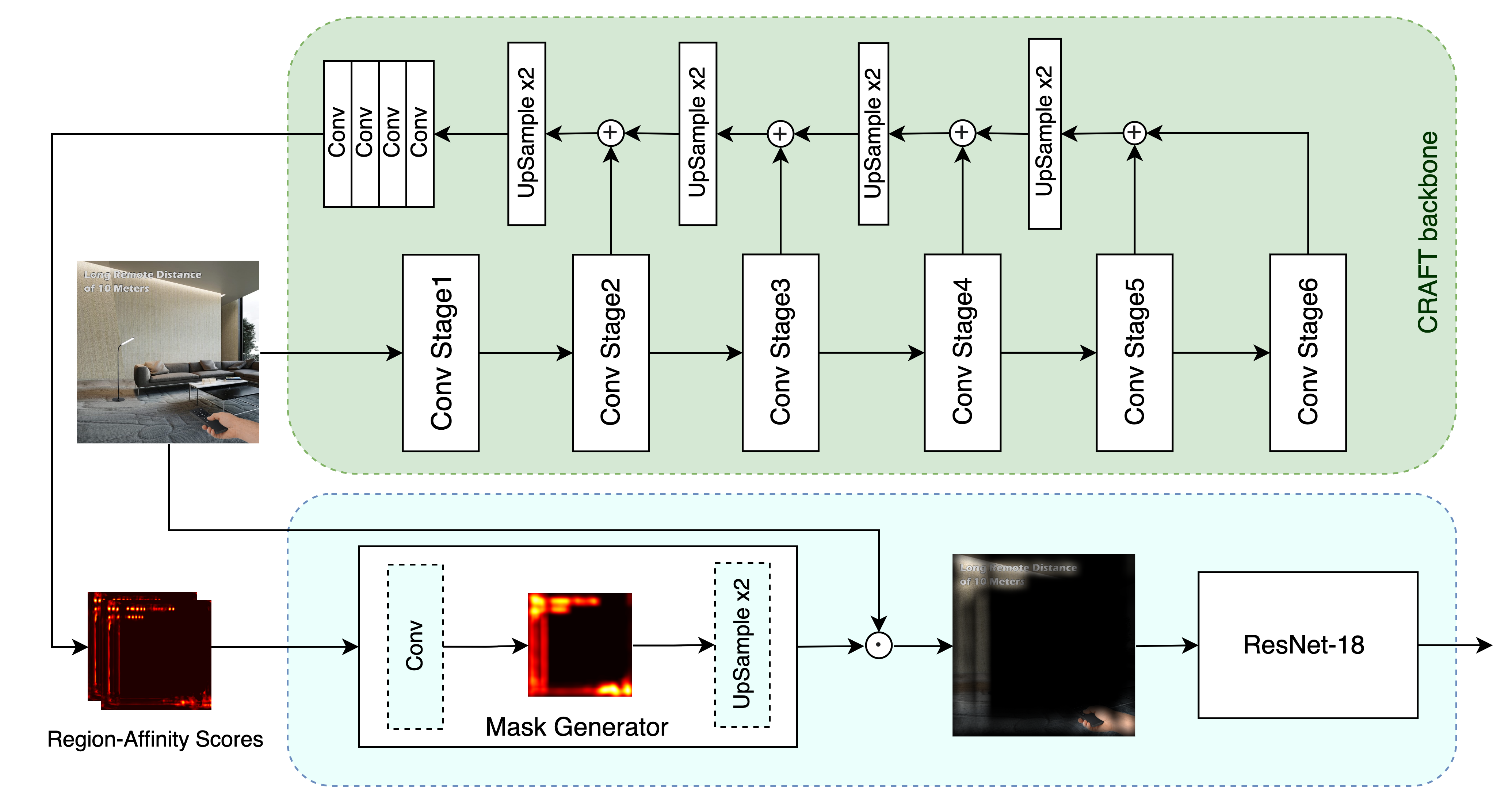}
 \caption{Diagram of the model architecture}
 \label{fig:model}
\end{figure}

We cast the problem of overlaying text detection as a binary image
classification task. For this purpose, all images with only scene text or no
text at all are considered as negative examples. Any image with a logo,
watermark, caption, or an overlaid text of any other form is considered to
belong to the positive class. This mapping corresponds to a practical scenario
of detecting defective images corrupted with unwanted text inscriptions.

Our overlaying text detection pipeline comprises of two stages. The first
component adapts the state-of-the-art text detector backbone for text
localization. On this stage, the CRAFT model scores are used to generate a
proposal mask. The second component utilizes the mask to attend to the regions
of interest in the input image, for subsequent classification a convolutional
neural network. Both components are trained together in an end-to-end manner.

The CRAFT model score is represented as a tensor $F^{\{\text{rg, af}\}}(x) \in
R^{\frac{h}{2} \times \frac{w}{2} \times 2}$ with the region score $rg$ and
affinity score $af$ channels:
\begin{equation}
  y = x \odot \underset{\times 2}{\operatorname{UpSample}} (H(F^{\{\text{rg,
      af}\}}(x))).
  \label{eq:model}
\end{equation}
The tensor $F^{\{\text{rg, af}\}}(x)$ is processed with a convolutional layer
$H$ with ReLU activation to produce a single channel $\frac{h}{2} \times
\frac{w}{2} \times 1$ mask. The resulting mask is upsampled with bilinear
interpolation by the factor of $2$ to match the dimensions, and combined with
the input $h \times w$ image $x$ with element-wise multiplication. Finally,
the masked image $y$ is fed into a ResNet-18~\cite{he2016deep} network for
classification.

The network architecture is schematically illustrated in
Figure~\ref{fig:model}.

We found that initializing weights of the convolutional layer $H$ with a
Gaussian kernel with a standard deviation of 8 pixels facilitates model
convergence. Gaussian blurring creates smoother information about text
neighborhoods, and also increases the size of the region where the text pixels
interact with the rest of the image pixels.

\section{Experiments}
\label{sec:experiments}
We compare the proposed CRAFT-masked ResNet approach with alternative
methods. As a simple baseline, we adapt the reference text detector approach
for binary classification task. To this end, the region and affinity score
maps from CRAFT model are flattened and fed into a linear binary classifier
algorithm. We refer to this method as ``Binarized CRAFT''. We also benchmark a
recent watermark and logo detection method SplitNet~\cite{cun2021split} for
overlaying text detection. Similarly to scene text detection, we build
``Binarized SplitNet'' and ``SplitNet-masked'' algorithms upon the watermark
mask matrix, extracted from a SplitNet model. The former method uses the
SplitNet mask directly as features for a binary classifier. Whereas the latter
leverages SplitNet backbone instead of CRAFT in our method to attend to
relevant image regions for classification with ResNet-18.

Finally, we perform ablation study to validate the impact of region attention
masks by training a vanilla ResNet-18 model directly on unmasked dataset
examples.

\subsection{Results}
We report overlaying text detection performance in binary classification
metrics, including precision, recall, the area under the ROC-curve (AUC), F1
score, as well as true positive and false positive rates (TPR and FPR,
correspondingly). Table~\ref{tab:results} summarizes performance on the
validation subset of Rooms-with-Text dataset among different detection
methods.

Our method outperforms alternative approaches in all metrics. We observe that
the watermark detection backbone pre-trained on synthetic data demonstrates an
inferior performance compared to text detection. Interestingly, we find that a
vanilla ResNet-18 network produces competitive results. However, its
significantly higher FPR and FNR make it impractical for real-world use cases.

\begin{table}[h]
  \resizebox{0.49\textwidth}{!}{%
  \centering
  \begin{tabular}{lllllll}
	\hline
	Method & AUC$\uparrow$ & Precision$\uparrow$ & Recall$\uparrow$ & F1 score$\uparrow$ & FPR$\downarrow$ & FNR$\downarrow$ \\
    \hline
	Binarized SplitNet & 0.70 & 0.59 & 0.43 & 0.53 & 0.20 & 0.53 \\
	SplitNet-masked    & 0.76 & 0.62 & 0.57 & 0.59 & 0.21 & 0.43 \\
	Binarized CRAFT    & 0.84 & 0.72 & 0.63 & 0.67 & 0.15 & 0.37 \\
	CRAFT-masked & \textbf{0.99} & \textbf{0.97} & \textbf{0.94} & \textbf{0.95} & \textbf{0.02} & \textbf{0.06} \\
    \hline
    ResNet-18          & 0.93 & 0.82 & 0.80 & 0.81 & 0.11 & 0.20 \\
  \hline
  \end{tabular}%
  }
  \caption{Overlaying text detection performance}
  \label{tab:results}
\end{table}

To better understand the failure modes, we provide qualitative results of the
proposed method in Appendix~\ref{sec:dataset_examples}.

\section{Conclusions}
\label{sec:conclusions}

In this work we proposed an annotated database of real-world interior images,
designed for the task of distinguishing overlaying text from scene text.

Motivated by characteristics of our dataset, we introduce a simple baseline
method for overlaying text detection. We validated and demonstrated the
efficiency of the character region-aware text detection backbone for
identifying relevant image areas for classification.

In future work, we plan to investigate recent advances in the field of image
statistics prior for extending the process from overlaying text detection to
image restoration and overlaying text removal.

{\small
\bibliographystyle{ieee_fullname}
\bibliography{rooms_with_text}

\begin{thebibliography}{10}\itemsep=-1pt

\bibitem{baek2019character}
Youngmin Baek, Bado Lee, Dongyoon Han, Sangdoo Yun, and Hwalsuk Lee.
\newblock Character region awareness for text detection.
\newblock In {\em Proceedings of the IEEE/CVF Conference on Computer Vision and
  Pattern Recognition}, pages 9365--9374, 2019.

\bibitem{bhardwaj2016image}
Dinesh Bhardwaj and Vinod Pankajakshan.
\newblock Image overlay text detection based on jpeg truncation error analysis.
\newblock {\em IEEE Signal Processing Letters}, 23(8):1027--1031, 2016.

\bibitem{ch2017total}
Chee~Kheng Ch'ng and Chee~Seng Chan.
\newblock Total-text: A comprehensive dataset for scene text detection and
  recognition.
\newblock In {\em 2017 14th IAPR international conference on document analysis
  and recognition (ICDAR)}, volume~1, pages 935--942. IEEE, 2017.

\bibitem{chng2019icdar2019}
Chee~Kheng Chng, Yuliang Liu, Yipeng Sun, Chun~Chet Ng, Canjie Luo, Zihan Ni,
  ChuanMing Fang, Shuaitao Zhang, Junyu Han, Errui Ding, et~al.
\newblock Icdar2019 robust reading challenge on arbitrary-shaped text-rrc-art.
\newblock In {\em 2019 International Conference on Document Analysis and
  Recognition (ICDAR)}, pages 1571--1576. IEEE, 2019.

\bibitem{cun2021split}
Xiaodong Cun and Chi-Man Pun.
\newblock Split then refine: stacked attention-guided resunets for blind single
  image visible watermark removal.
\newblock In {\em Proceedings of the AAAI Conference on Artificial
  Intelligence}, volume~35, pages 1184--1192, 2021.

\bibitem{dashti2015video}
Maryam Dashti, Reza Safabakhsh, Mohammadreza Pourfard, and Mohammadjavad
  Abdollahifard.
\newblock Video logo removal using iterative subsequent matching.
\newblock In {\em 2015 The International Symposium on Artificial Intelligence
  and Signal Processing (AISP)}, pages 84--88. IEEE, 2015.

\bibitem{dekel2017effectiveness}
Tali Dekel, Michael Rubinstein, Ce Liu, and William~T Freeman.
\newblock On the effectiveness of visible watermarks.
\newblock In {\em Proceedings of the IEEE Conference on Computer Vision and
  Pattern Recognition}, pages 2146--2154, 2017.

\bibitem{deng2018pixellink}
Dan Deng, Haifeng Liu, Xuelong Li, and Deng Cai.
\newblock Pixellink: Detecting scene text via instance segmentation.
\newblock In {\em Proceedings of the AAAI Conference on Artificial
  Intelligence}, volume~32, 2018.

\bibitem{gllavata2004text}
Julinda Gllavata, Ralph Ewerth, and Bernd Freisleben.
\newblock Text detection in images based on unsupervised classification of
  high-frequency wavelet coefficients.
\newblock In {\em Proceedings of the 17th International Conference on Pattern
  Recognition, 2004. ICPR 2004.}, volume~1, pages 425--428. IEEE, 2004.

\bibitem{gupta2016synthetic}
Ankush Gupta, Andrea Vedaldi, and Andrew Zisserman.
\newblock Synthetic data for text localisation in natural images.
\newblock In {\em Proceedings of the IEEE conference on computer vision and
  pattern recognition}, pages 2315--2324, 2016.

\bibitem{he2016deep}
Kaiming He, Xiangyu Zhang, Shaoqing Ren, and Jian Sun.
\newblock Deep residual learning for image recognition.
\newblock In {\em Proceedings of the IEEE conference on computer vision and
  pattern recognition}, pages 770--778, 2016.

\bibitem{hoi2015logo}
Steven~CH Hoi, Xiongwei Wu, Hantang Liu, Yue Wu, Huiqiong Wang, Hui Xue, and
  Qiang Wu.
\newblock Logo-net: Large-scale deep logo detection and brand recognition with
  deep region-based convolutional networks.
\newblock {\em arXiv preprint arXiv:1511.02462}, 2015.

\bibitem{jain2021logonet}
Rahul~Kumar Jain, Taro Watasue, Tomohiro Nakagawa, SATO Takahiro, Yutaro
  Iwamoto, RUAN Xiang, and CHEN Yen-Wei.
\newblock Logonet: Layer-aggregated attention centernet for logo detection.
\newblock In {\em 2021 IEEE International Conference on Consumer Electronics
  (ICCE)}, pages 1--6. IEEE, 2021.

\bibitem{joly2009logo}
Alexis Joly and Olivier Buisson.
\newblock Logo retrieval with a contrario visual query expansion.
\newblock In {\em Proceedings of the 17th ACM international conference on
  Multimedia}, pages 581--584, 2009.

\bibitem{karatzas2015icdar}
Dimosthenis Karatzas, Lluis Gomez-Bigorda, Anguelos Nicolaou, Suman Ghosh,
  Andrew Bagdanov, Masakazu Iwamura, Jiri Matas, Lukas Neumann,
  Vijay~Ramaseshan Chandrasekhar, Shijian Lu, et~al.
\newblock Icdar 2015 competition on robust reading.
\newblock In {\em 2015 13th international conference on document analysis and
  recognition (ICDAR)}, pages 1156--1160. IEEE, 2015.

\bibitem{karatzas2013icdar}
Dimosthenis Karatzas, Faisal Shafait, Seiichi Uchida, Masakazu Iwamura,
  Lluis~Gomez i Bigorda, Sergi~Robles Mestre, Joan Mas, David~Fernandez Mota,
  Jon~Almazan Almazan, and Lluis~Pere De~Las~Heras.
\newblock Icdar 2013 robust reading competition.
\newblock In {\em 2013 12th International Conference on Document Analysis and
  Recognition}, pages 1484--1493. IEEE, 2013.

\bibitem{kim2008new}
Wonjun Kim and Changick Kim.
\newblock A new approach for overlay text detection and extraction from complex
  video scene.
\newblock {\em IEEE transactions on image processing}, 18(2):401--411, 2008.

\bibitem{kumari2018three}
Lalita Kumari, Vidyut Dey, and JL Raheja.
\newblock A three-layer approach for overlay text extraction in video stream.
\newblock In {\em Soft Computing: Theories and Applications}, pages 79--87.
  Springer, 2018.

\bibitem{li2020system}
Liam Li, Kevin Jamieson, Afshin Rostamizadeh, Ekaterina Gonina, Jonathan
  Ben-Tzur, Moritz Hardt, Benjamin Recht, and Ameet Talwalkar.
\newblock A system for massively parallel hyperparameter tuning.
\newblock {\em Proceedings of Machine Learning and Systems}, 2:230--246, 2020.

\bibitem{li2010fast}
Zhe Li, Matthias Schulte-Austum, and Martin Neschen.
\newblock Fast logo detection and recognition in document images.
\newblock In {\em 2010 20th International Conference on Pattern Recognition},
  pages 2716--2719. IEEE, 2010.

\bibitem{liao2017textboxes}
Minghui Liao, Baoguang Shi, Xiang Bai, Xinggang Wang, and Wenyu Liu.
\newblock Textboxes: A fast text detector with a single deep neural network.
\newblock In {\em Thirty-first AAAI conference on artificial intelligence},
  2017.

\bibitem{lienhart2002localizing}
Rainer Lienhart and Axel Wernicke.
\newblock Localizing and segmenting text in images and videos.
\newblock {\em IEEE Transactions on circuits and systems for video technology},
  12(4):256--268, 2002.

\bibitem{lin2014microsoft}
Tsung-Yi Lin, Michael Maire, Serge Belongie, James Hays, Pietro Perona, Deva
  Ramanan, Piotr Doll{\'a}r, and C~Lawrence Zitnick.
\newblock Microsoft coco: Common objects in context.
\newblock In {\em European conference on computer vision}, pages 740--755.
  Springer, 2014.

\bibitem{liu2018fots}
Xuebo Liu, Ding Liang, Shi Yan, Dagui Chen, Yu Qiao, and Junjie Yan.
\newblock Fots: Fast oriented text spotting with a unified network.
\newblock In {\em Proceedings of the IEEE conference on computer vision and
  pattern recognition}, pages 5676--5685, 2018.

\bibitem{nayef2017icdar2017}
Nibal Nayef, Fei Yin, Imen Bizid, Hyunsoo Choi, Yuan Feng, Dimosthenis
  Karatzas, Zhenbo Luo, Umapada Pal, Christophe Rigaud, Joseph Chazalon, et~al.
\newblock Icdar2017 robust reading challenge on multi-lingual scene text
  detection and script identification.
\newblock In {\em 2017 14th IAPR International Conference on Document Analysis
  and Recognition (ICDAR)}, volume~1, pages 1454--1459. IEEE, 2017.

\bibitem{paszke2019pytorch}
Adam Paszke, Sam Gross, Francisco Massa, Adam Lerer, James Bradbury, Gregory
  Chanan, Trevor Killeen, Zeming Lin, Natalia Gimelshein, Luca Antiga, et~al.
\newblock Pytorch: An imperative style, high-performance deep learning library.
\newblock {\em Advances in neural information processing systems}, 32, 2019.

\bibitem{quehl2015improving}
Bernhard Quehl, Haojin Yang, and Harald Sack.
\newblock Improving text recognition by distinguishing scene and overlay text.
\newblock In {\em Seventh International Conference on Machine Vision (ICMV
  2014)}, volume 9445, page 944509. International Society for Optics and
  Photonics, 2015.

\bibitem{romberg2011scalable}
Stefan Romberg, Lluis~Garcia Pueyo, Rainer Lienhart, and Roelof Van~Zwol.
\newblock Scalable logo recognition in real-world images.
\newblock In {\em Proceedings of the 1st ACM International Conference on
  Multimedia Retrieval}, pages 1--8, 2011.

\bibitem{sage2018logo}
Alexander Sage, Eirikur Agustsson, Radu Timofte, and Luc Van~Gool.
\newblock Logo synthesis and manipulation with clustered generative adversarial
  networks.
\newblock In {\em Proceedings of the IEEE Conference on Computer Vision and
  Pattern Recognition}, pages 5879--5888, 2018.

\bibitem{shen2021large}
Xi Shen, Ilaria Pastrolin, Oumayma Bounou, Spyros Gidaris, Marc Smith, Olivier
  Poncet, and Mathieu Aubry.
\newblock Large-scale historical watermark recognition: dataset and a new
  consistency-based approach.
\newblock In {\em 2020 25th International Conference on Pattern Recognition
  (ICPR)}, pages 6810--6817. IEEE, 2021.

\bibitem{shivakumara2014separation}
P Shivakumara, N~Vinay Kumar, DS Guru, and Chew~Lim Tan.
\newblock Separation of graphics (superimposed) and scene text in video frames.
\newblock In {\em 2014 11th IAPR International Workshop on Document Analysis
  Systems}, pages 344--348. IEEE, 2014.

\bibitem{slucki2018extracting}
Adam S{\l}ucki, Tomasz Trzci{\'n}ski, Adam Bielski, and Pawe{\l} Cyrta.
\newblock Extracting textual overlays from social media videos using neural
  networks.
\newblock In {\em International Conference on Computer Vision and Graphics},
  pages 287--299. Springer, 2018.

\bibitem{su2017deep}
Hang Su, Xiatian Zhu, and Shaogang Gong.
\newblock Deep learning logo detection with data expansion by synthesising
  context.
\newblock In {\em 2017 IEEE Winter Conference on Applications of Computer
  Vision (WACV)}, pages 530--539. IEEE Computer Society, 2017.

\bibitem{ulyanov2018deep}
Dmitry Ulyanov, Andrea Vedaldi, and Victor Lempitsky.
\newblock Deep image prior.
\newblock In {\em Proceedings of the IEEE conference on computer vision and
  pattern recognition}, pages 9446--9454, 2018.

\bibitem{veit2016coco}
Andreas Veit, Tomas Matera, Lukas Neumann, Jiri Matas, and Serge Belongie.
\newblock Coco-text: Dataset and benchmark for text detection and recognition
  in natural images.
\newblock {\em arXiv preprint arXiv:1601.07140}, 2016.

\bibitem{wang2007automatic}
Jinqiao Wang, Qingshan Liu, Lingyu Duan, Hanqing Lu, and Changsheng Xu.
\newblock Automatic tv logo detection, tracking and removal in broadcast video.
\newblock In {\em International Conference on Multimedia Modeling}, pages
  63--72. Springer, 2007.

\bibitem{wang2022logodet}
Jing Wang, Weiqing Min, Sujuan Hou, Shengnan Ma, Yuanjie Zheng, and Shuqiang
  Jiang.
\newblock Logodet-3k: A large-scale image dataset for logo detection.
\newblock {\em ACM Transactions on Multimedia Computing, Communications, and
  Applications (TOMM)}, 18(1):1--19, 2022.

\bibitem{yan2005automatic}
Wei-Qi Yan, Jun Wang, and Mohan~S Kankanhalli.
\newblock Automatic video logo detection and removal.
\newblock {\em Multimedia Systems}, 10(5):379--391, 2005.

\bibitem{yao2012detecting}
Cong Yao, Xiang Bai, Wenyu Liu, Yi Ma, and Zhuowen Tu.
\newblock Detecting texts of arbitrary orientations in natural images.
\newblock In {\em 2012 IEEE conference on computer vision and pattern
  recognition}, pages 1083--1090. IEEE, 2012.

\bibitem{zhang2003accurate}
DongQing Zhang, Belle~L Tseng, and S-F Chang.
\newblock Accurate overlay text extraction for digital video analysis.
\newblock In {\em International Conference on Information Technology: Research
  and Education, 2003. Proceedings. ITRE2003.}, pages 233--237. IEEE, 2003.

\bibitem{zhou2017east}
Xinyu Zhou, Cong Yao, He Wen, Yuzhi Wang, Shuchang Zhou, Weiran He, and Jiajun
  Liang.
\newblock East: an efficient and accurate scene text detector.
\newblock In {\em Proceedings of the IEEE conference on Computer Vision and
  Pattern Recognition}, pages 5551--5560, 2017.

\end{thebibliography}
}

\appendix

\section{Appendix: Experimental Setup}

We use PyTorch~\cite{paszke2019pytorch} framework to implement our models. We
perform training and evaluation on Rooms-with-Text dataset. Examples with
labels ``Overlaying'' and ``Both'' were mapped to the positive class, whereas
examples labeled as ``Scene'' and ``None'' were considered as the negative
class. The input images were resized and padded to the dimensions as required
by corresponding models with aspect ratio preservation.

We use a mini-batch size of 32, and stochastic gradient descent optimizer with
an initial learning rate of 0.015 and momentum 0.9. The learning rate was
annealed by the factor of 0.5 after each epoch, when a plateau was
encountered. The optimal combination of hyperparameters was found by tuning on
the validation set with the asynchronous Hyperband~\cite{li2020system}
algorithm.

All networks were trained until convergence with the binary cross-entropy
objective. To combat overfitting, a weight decay penalty of $1 \times 10^{-5}$
was added to all model parameters. We have not used any additional data
augmentation techniques during model training.

Following the training protocol in~\cite{cun2021split}, SplitNet backbone was
pre-trained on synthesized datasets derived from background images from
MSCOCO~\cite{lin2014microsoft}. The CRAFT backbone was pre-trained on
SynthText~\cite{gupta2016synthetic}, and then fine-tuned on
ICDAR2013~\cite{karatzas2013icdar} and ICDAR2017~\cite{nayef2017icdar2017}.

\section{Appendix: Dataset Examples}
\label{sec:dataset_examples}

\begin{figure*}[h]
  \centering
  \begin{subfigure}[b]{0.19\textwidth}
    \centering
    \includegraphics[width=\textwidth]{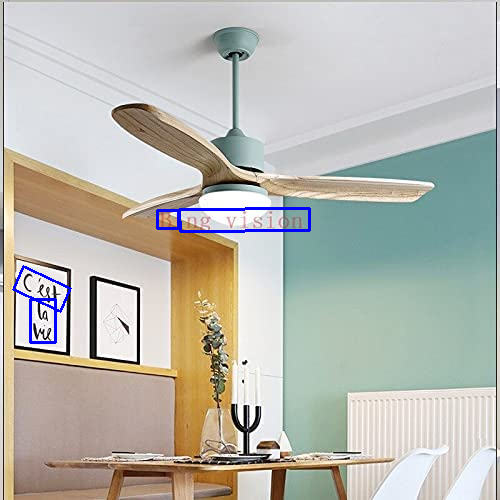}
  \end{subfigure}
  \begin{subfigure}[b]{0.19\textwidth}
    \centering
    \includegraphics[width=\textwidth]{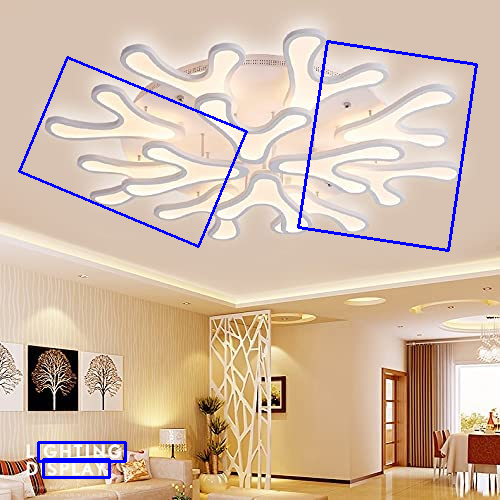}
  \end{subfigure}
  \begin{subfigure}[b]{0.19\textwidth}
    \centering
    \includegraphics[width=\textwidth]{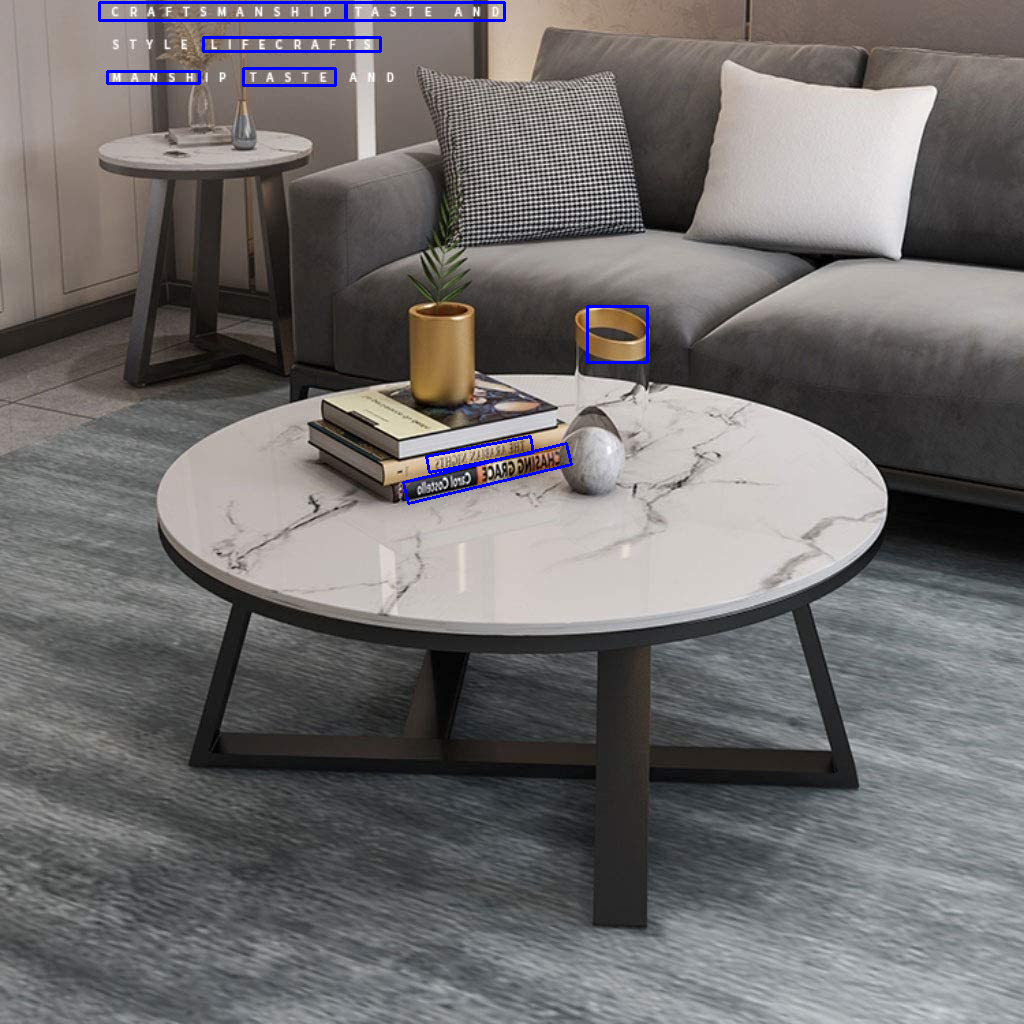}
  \end{subfigure}
  \begin{subfigure}[b]{0.19\textwidth}
    \centering
    \includegraphics[width=\textwidth]{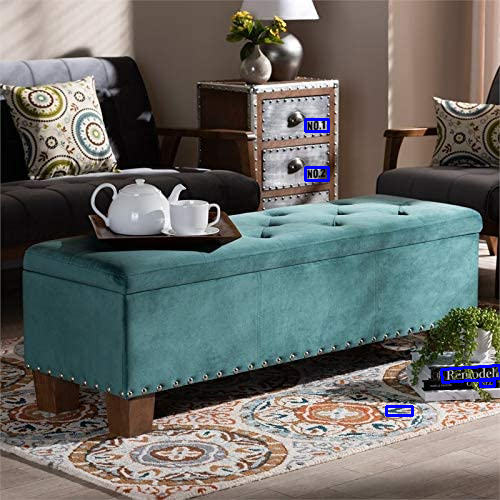}
  \end{subfigure}
  \begin{subfigure}[b]{0.19\textwidth}
    \centering
    \includegraphics[width=\textwidth]{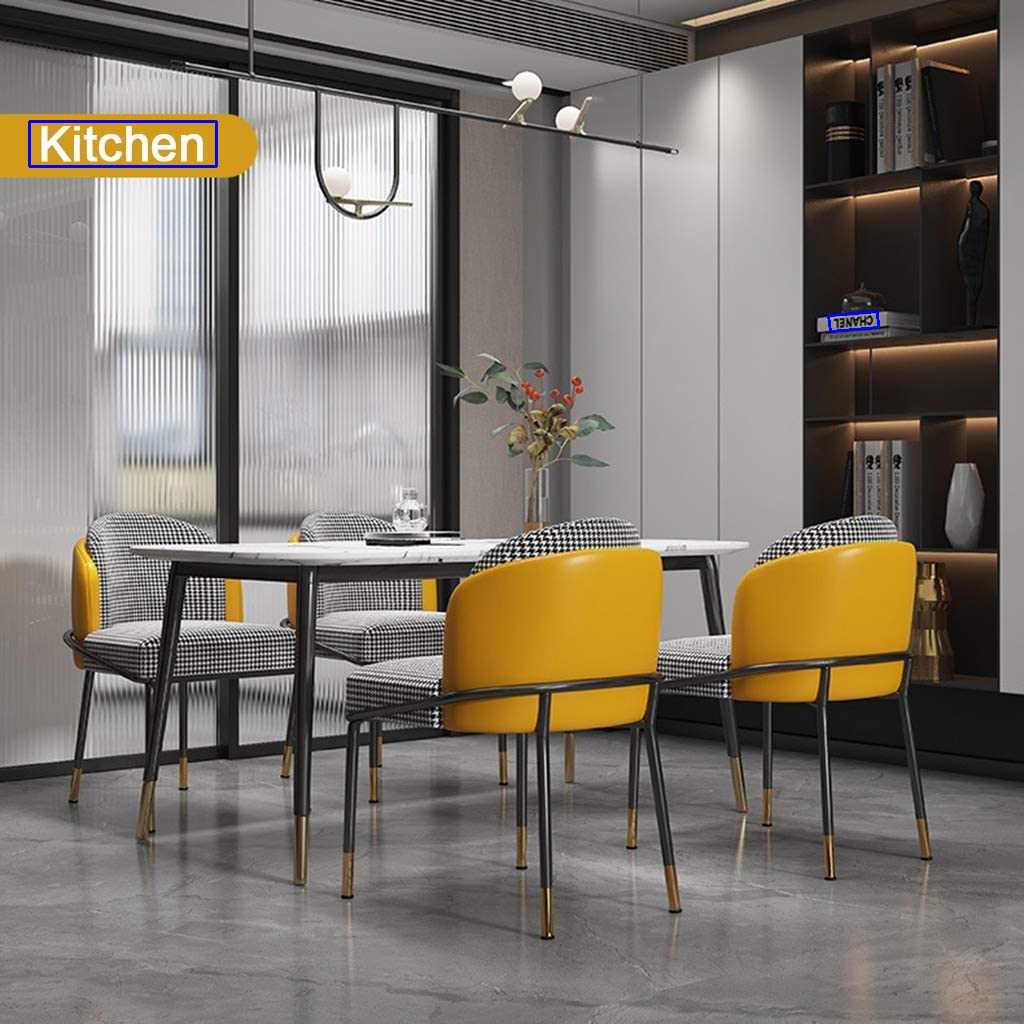}
  \end{subfigure}
  \begin{subfigure}[b]{0.19\textwidth}
    \centering
    \includegraphics[width=\textwidth]{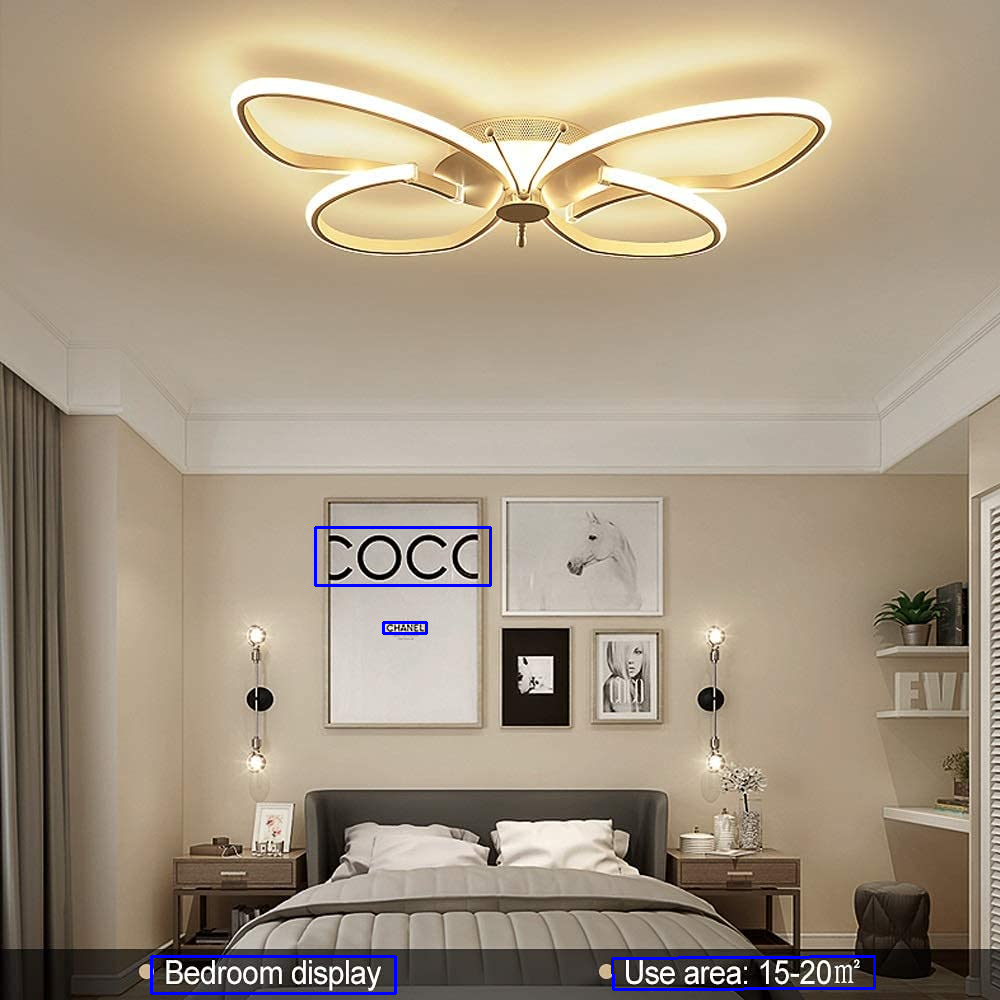}
  \end{subfigure}
  \begin{subfigure}[b]{0.19\textwidth}
    \centering
    \includegraphics[width=\textwidth]{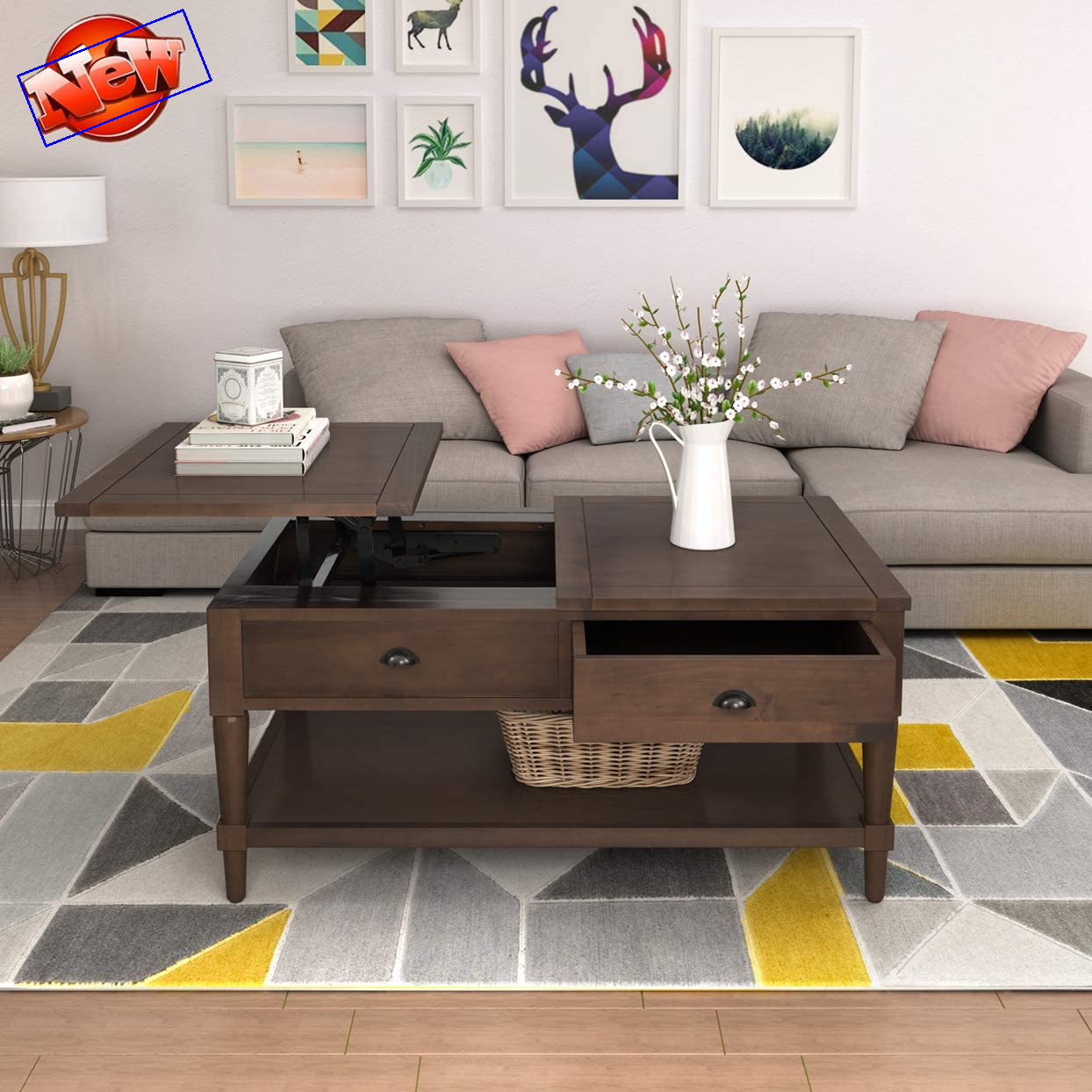}
  \end{subfigure}
  \begin{subfigure}[b]{0.19\textwidth}
    \centering
    \includegraphics[width=\textwidth]{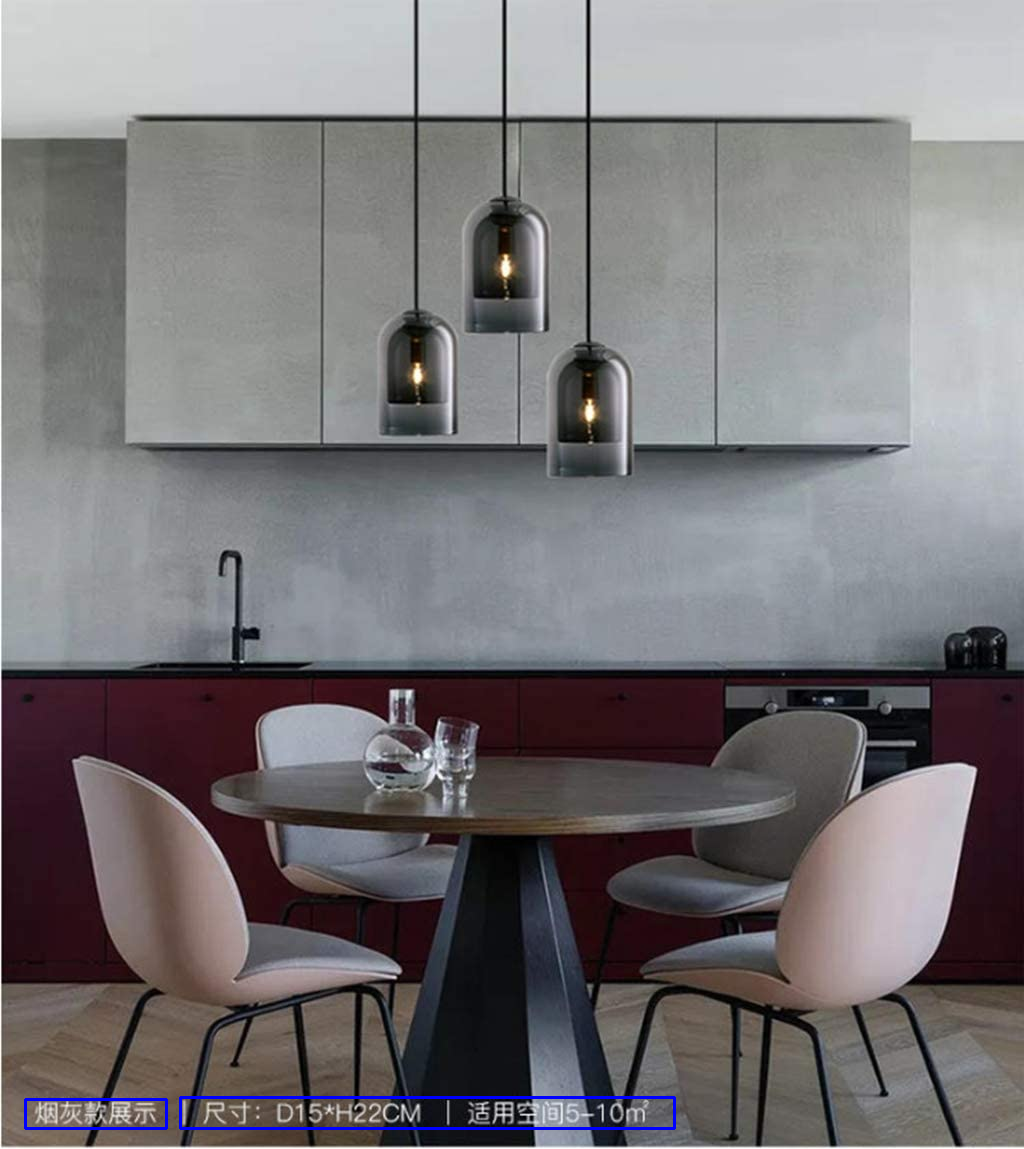}
  \end{subfigure}
  \begin{subfigure}[b]{0.19\textwidth}
    \centering
    \includegraphics[width=\textwidth]{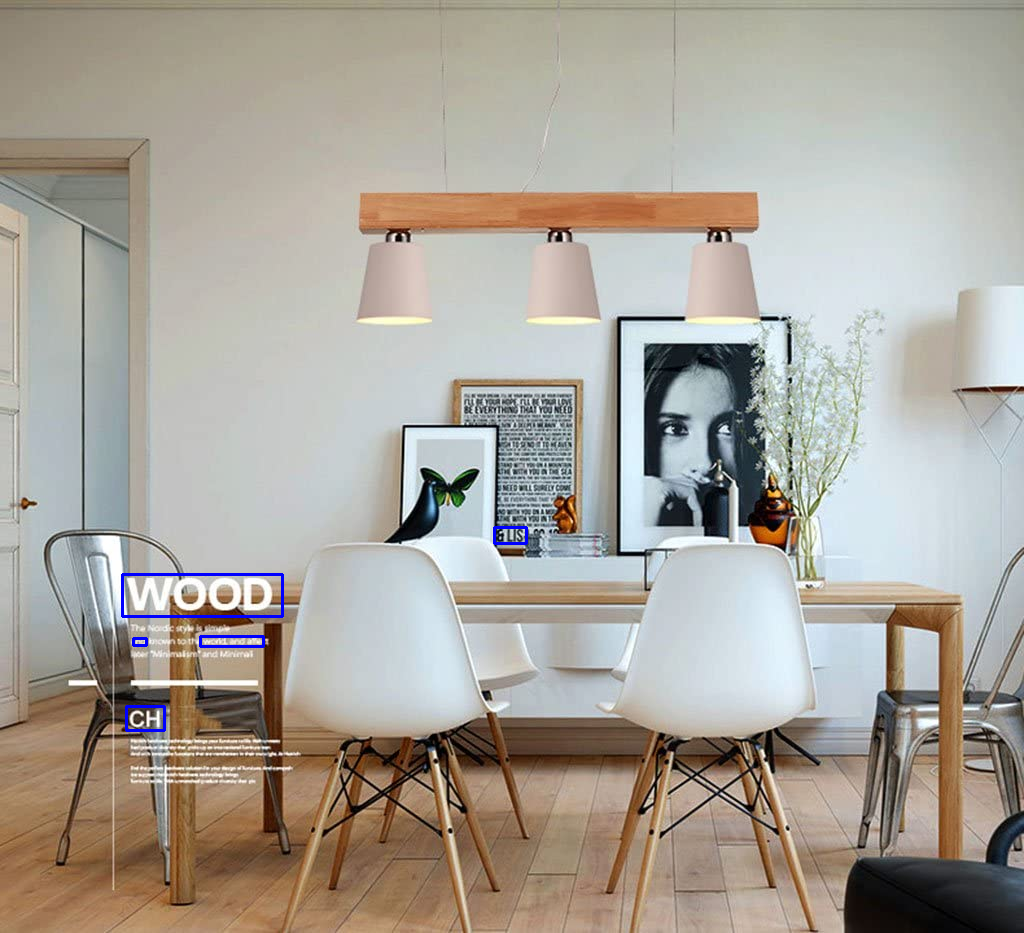}
  \end{subfigure}
  \begin{subfigure}[b]{0.19\textwidth}
    \centering
    \includegraphics[width=\textwidth]{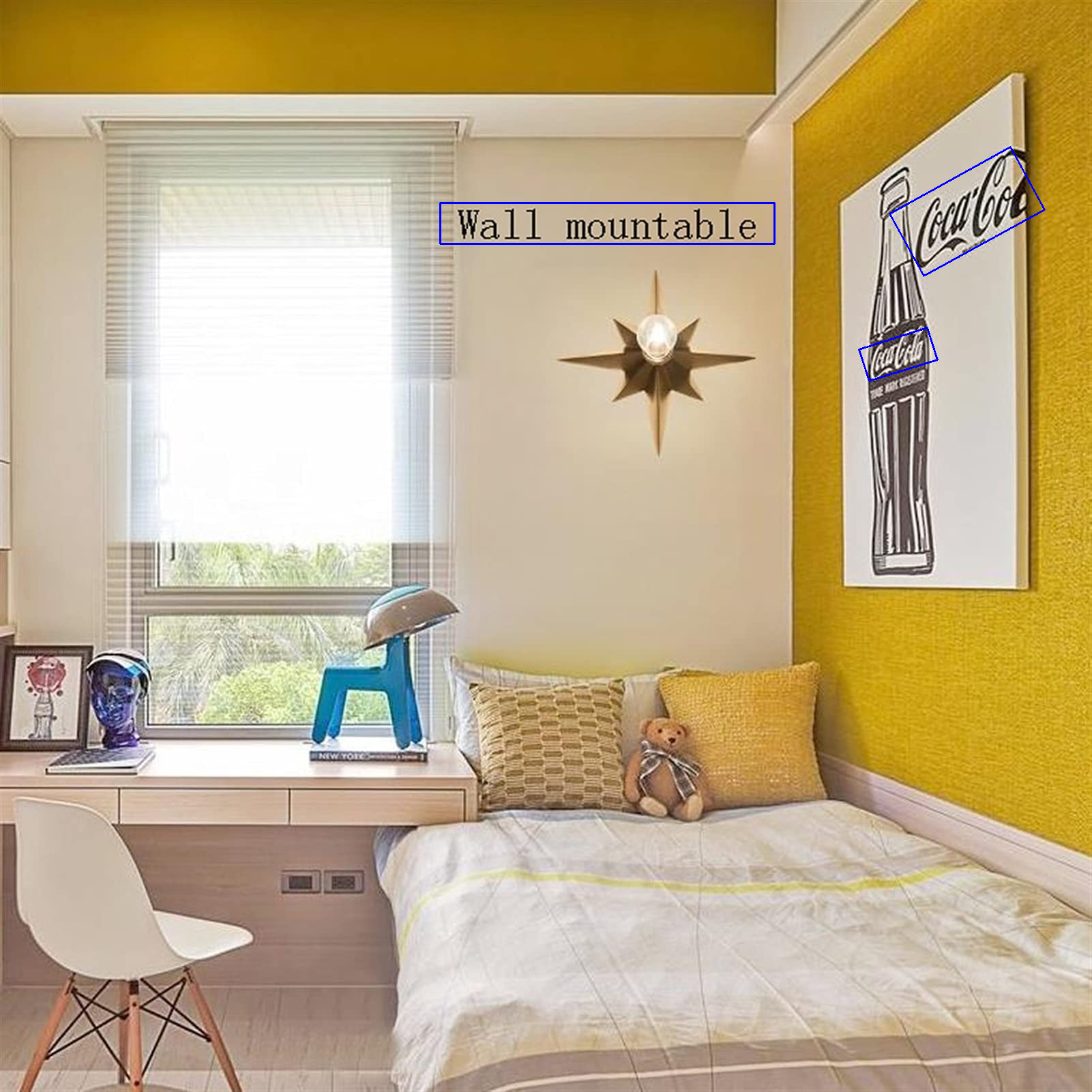}
  \end{subfigure}
  \caption{Examples of real-world product images with scene and overlaying
    text. Bounding boxes are generated with the CRAFT~\cite{baek2019character}
    text detector for illustration purposes. Best viewed on screen}
  \label{fig:dataset_examples}
\end{figure*}

\begin{figure*}[h]
  \centering
  \begin{subfigure}[b]{0.19\textwidth}
    \centering
    \includegraphics[width=\textwidth]{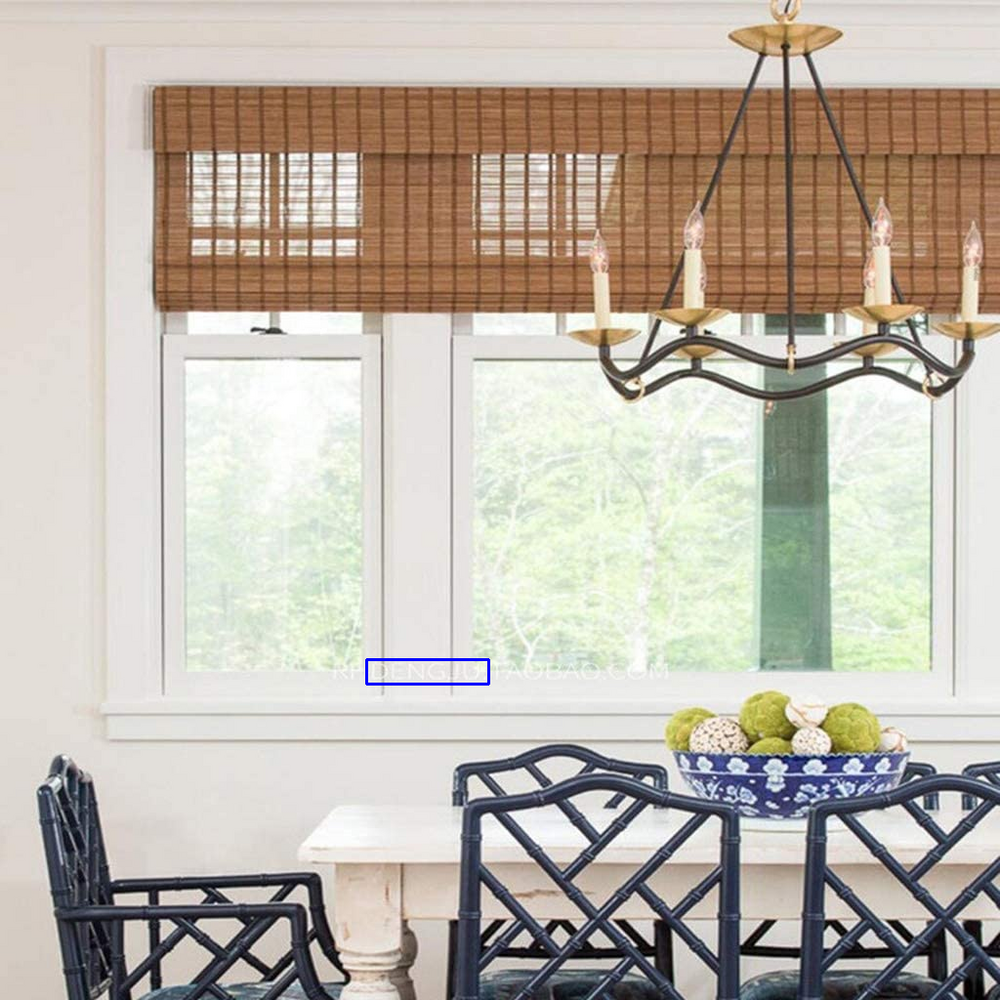}
  \end{subfigure}
  \begin{subfigure}[b]{0.19\textwidth}
    \centering
    \includegraphics[width=\textwidth]{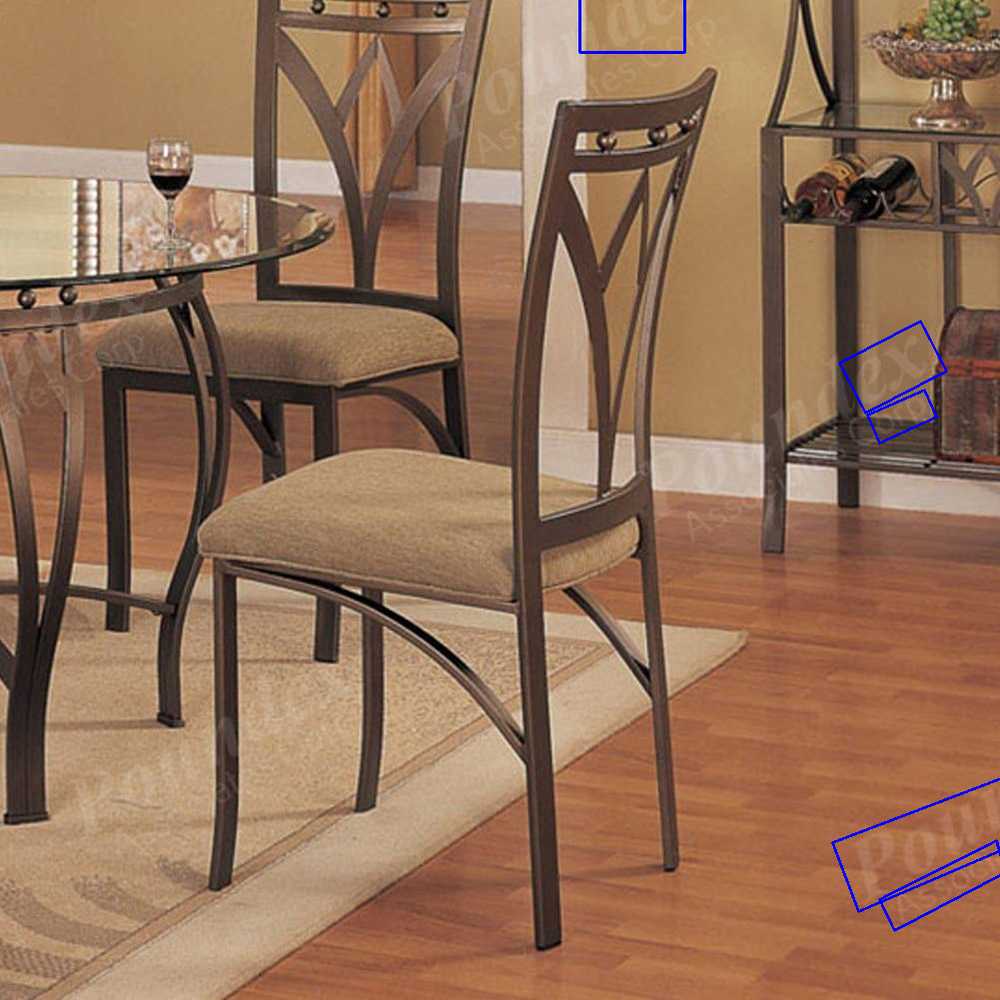}
  \end{subfigure}
  \begin{subfigure}[b]{0.19\textwidth}
    \centering
    \includegraphics[width=\textwidth]{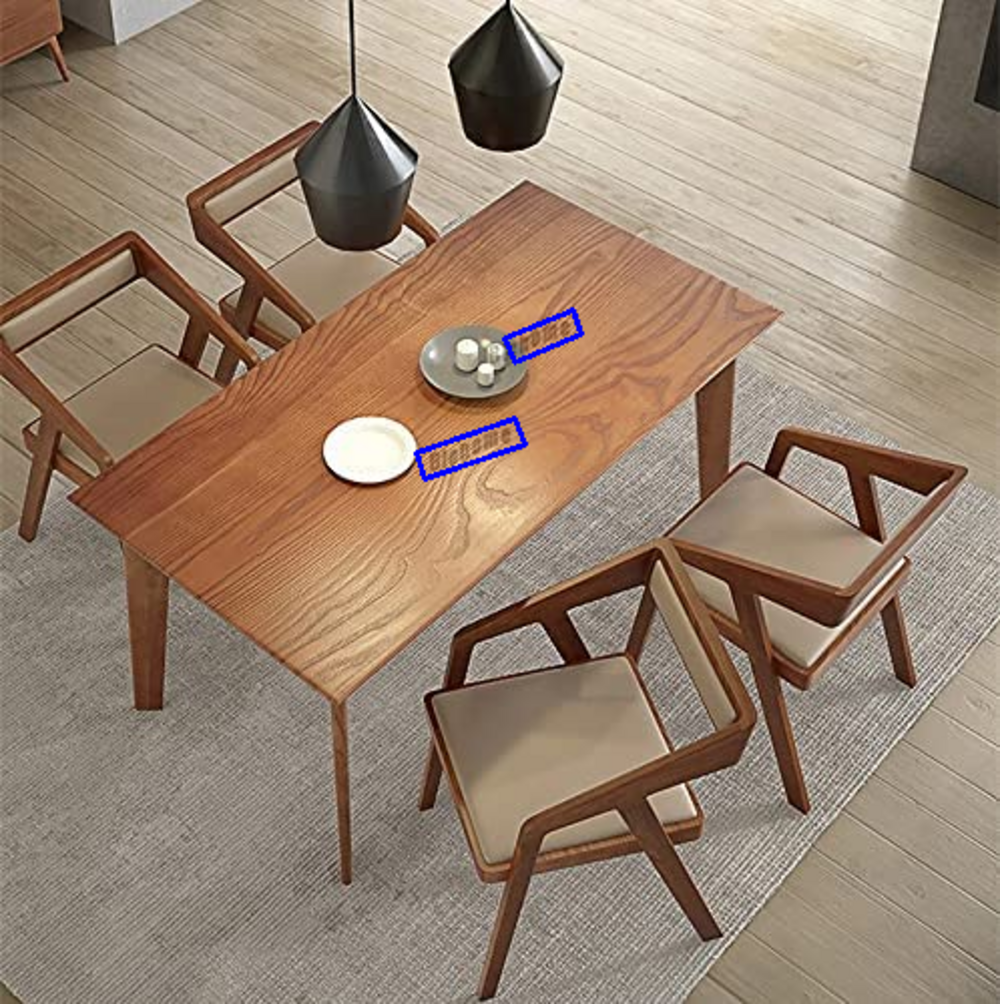}
  \end{subfigure}
  \begin{subfigure}[b]{0.19\textwidth}
    \centering
    \includegraphics[width=\textwidth]{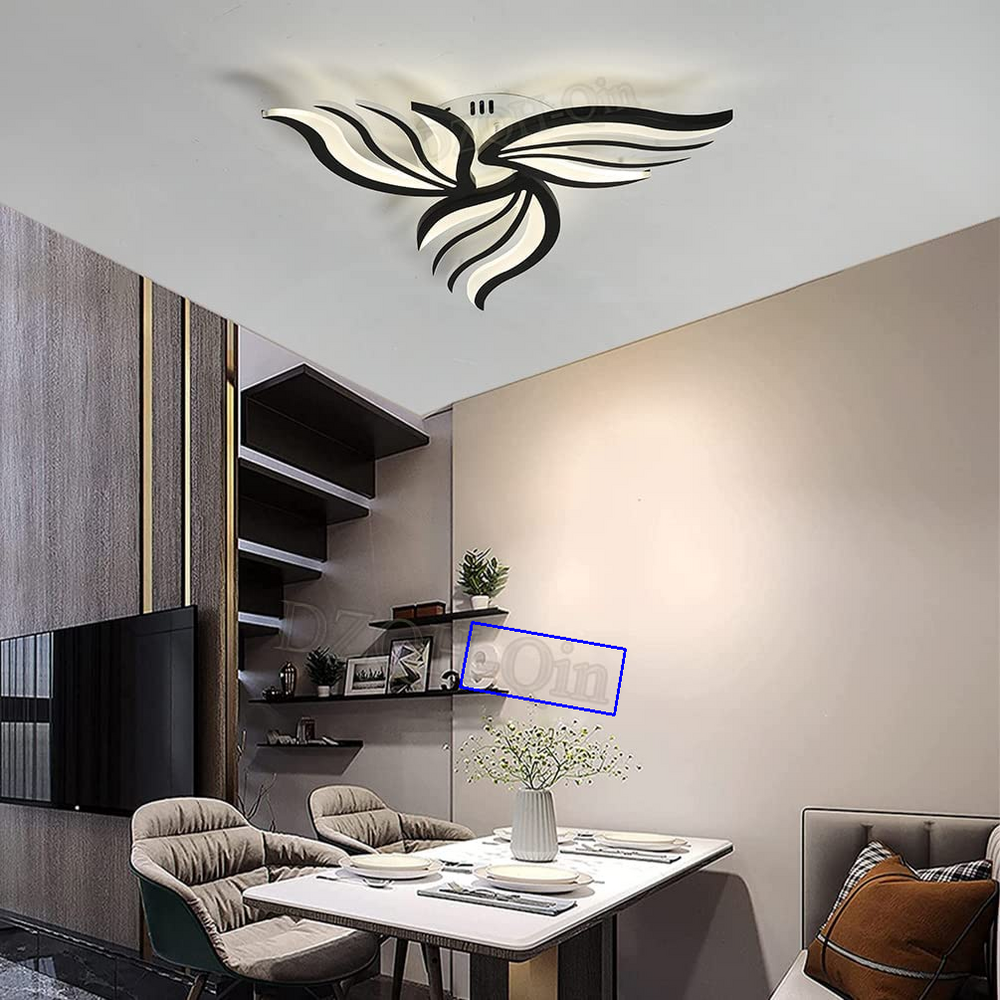}
  \end{subfigure}
  \begin{subfigure}[b]{0.19\textwidth}
    \centering
    \includegraphics[width=\textwidth]{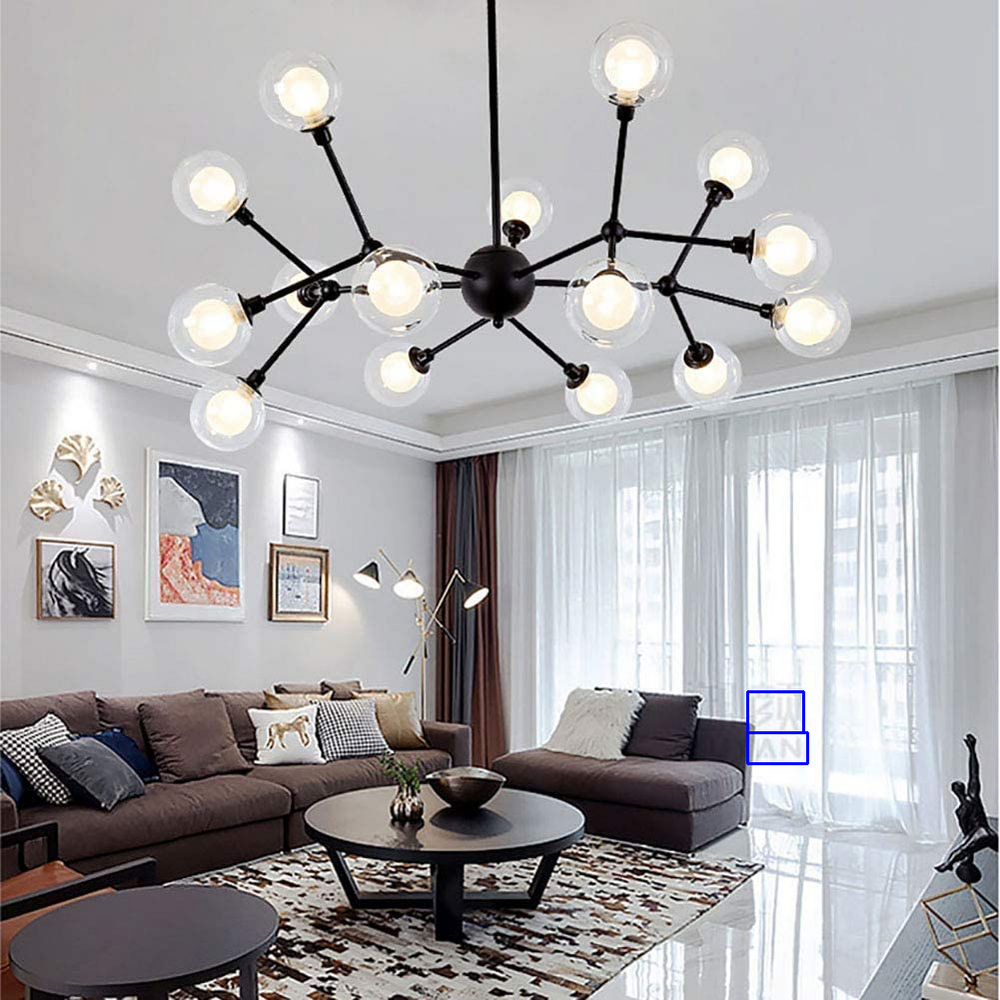}
  \end{subfigure}
  \begin{subfigure}[b]{0.19\textwidth}
    \centering
    \includegraphics[width=\textwidth]{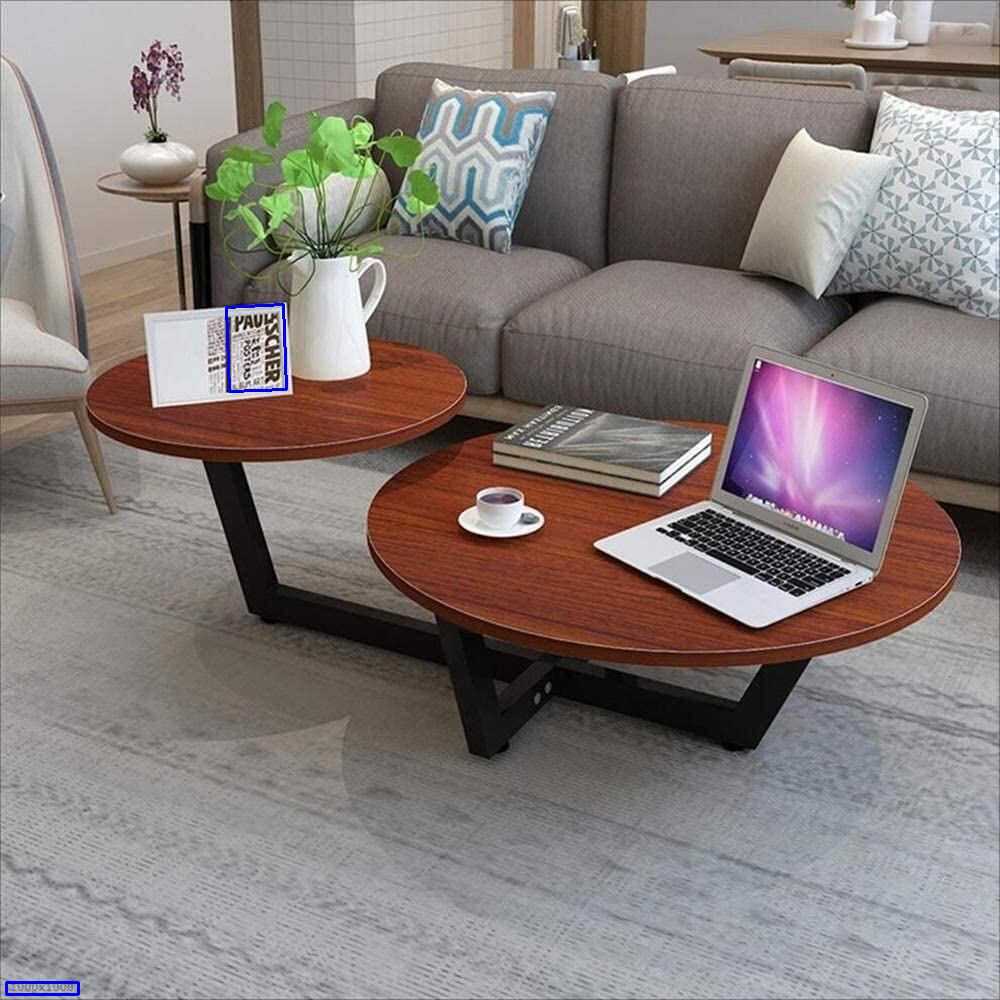}
  \end{subfigure}
  \begin{subfigure}[b]{0.19\textwidth}
    \centering
    \includegraphics[width=\textwidth]{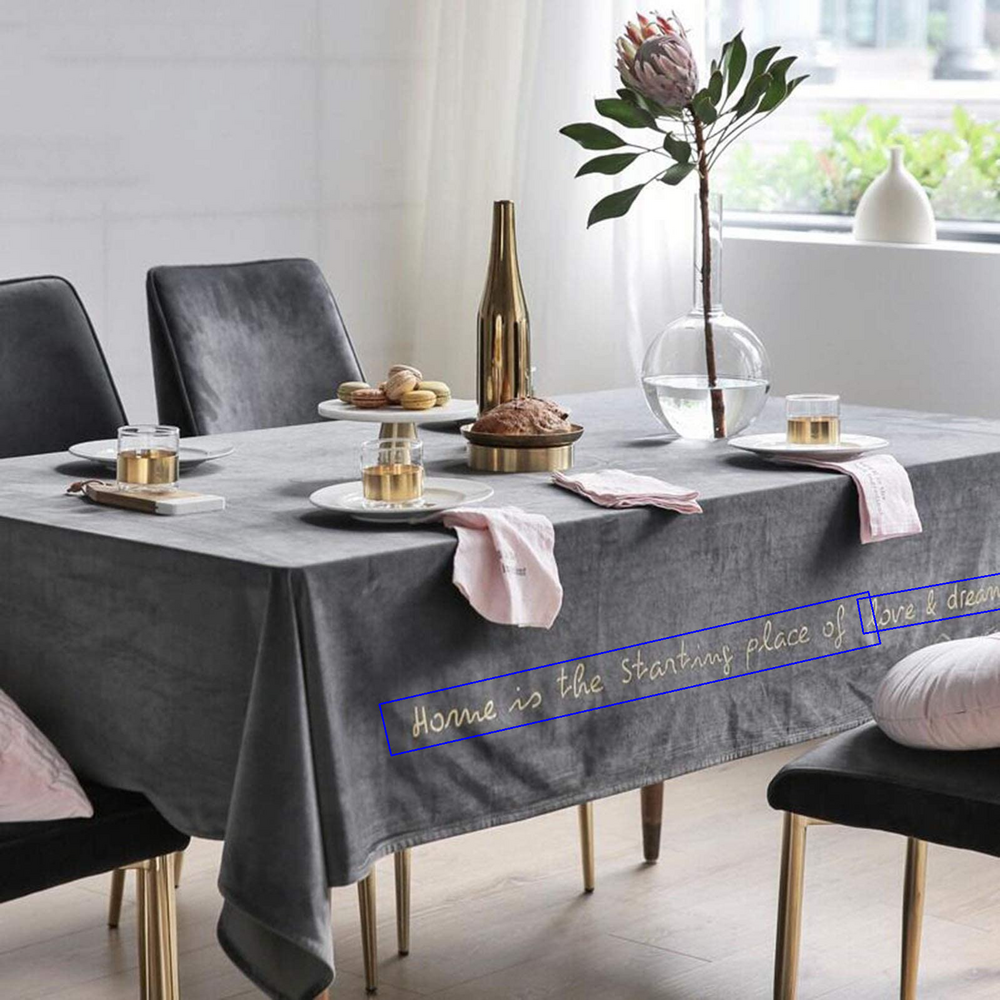}
  \end{subfigure}
  \begin{subfigure}[b]{0.19\textwidth}
    \centering
    \includegraphics[width=\textwidth]{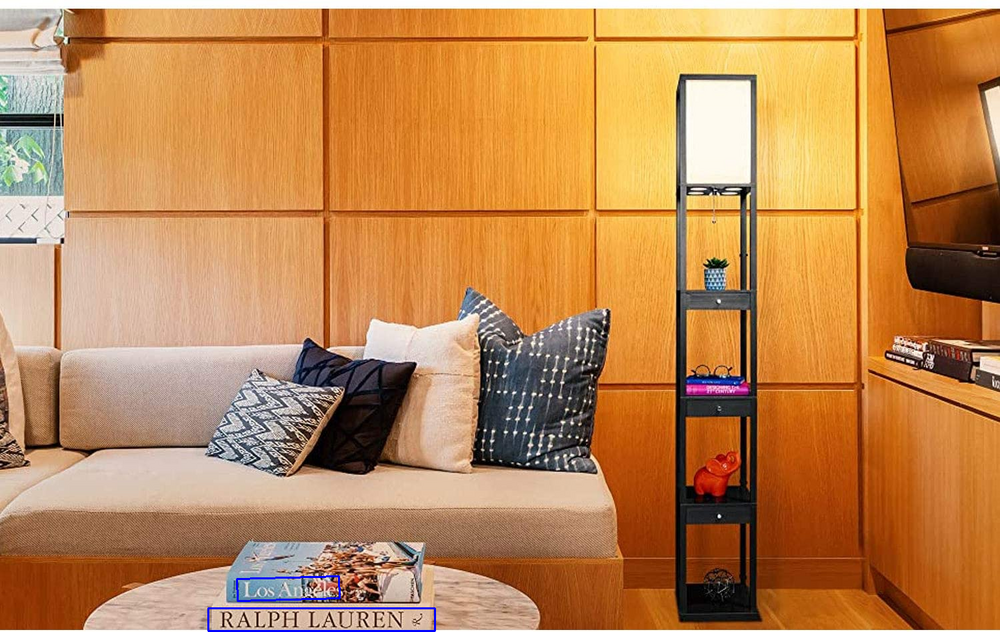}
  \end{subfigure}
  \begin{subfigure}[b]{0.19\textwidth}
    \centering
    \includegraphics[width=\textwidth]{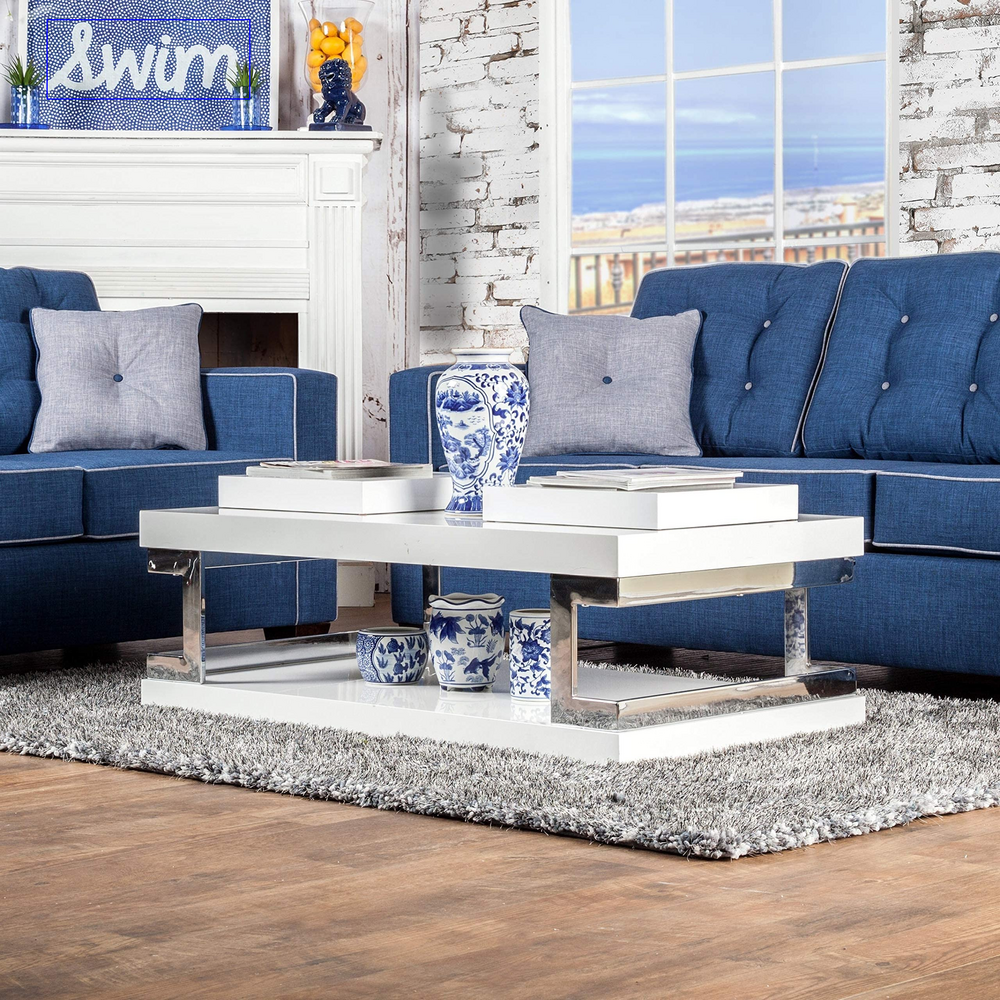}
  \end{subfigure}
  \begin{subfigure}[b]{0.19\textwidth}
    \centering
    \includegraphics[width=\textwidth]{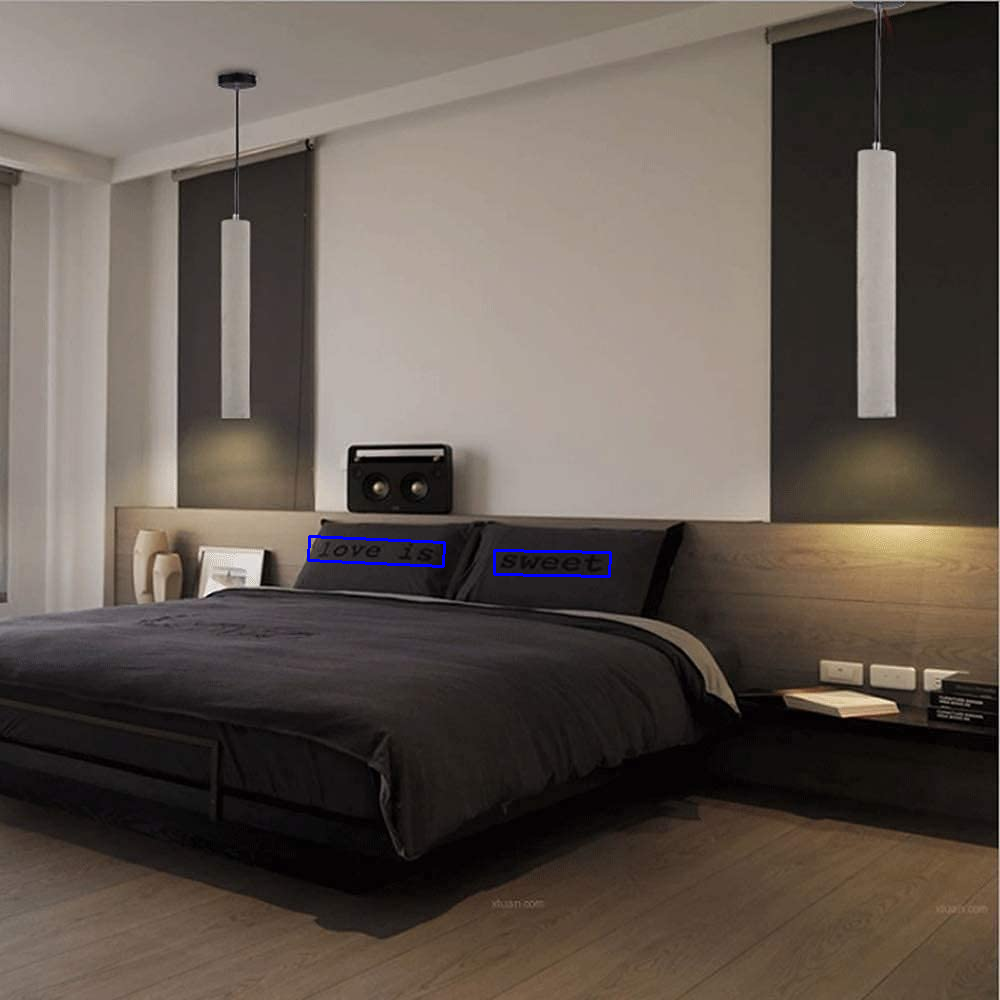}
  \end{subfigure}
  \caption{Qualitative results of the proposed method. The top row and the
    bottom row depict false negative and false positive predictions
    correspondingly, along with bounding boxes generated by the
    CRAFT~\cite{baek2019character} text detector. Best viewed on screen}
  \label{fig:qualitative}
\end{figure*}

\end{document}